\documentclass[letterpaper]{article} 
\usepackage{aaai24}  
\usepackage{times}  
\usepackage{helvet}  
\usepackage{courier}  
\usepackage[hyphens]{url}  
\usepackage{graphicx} 
\usepackage{natbib}  
\usepackage{caption} 
\usepackage{algorithm}
\usepackage{algorithmic}

\usepackage{pdfpages}
\usepackage{xurl}
\usepackage{amsmath}
\usepackage{amsthm}
\usepackage{booktabs}
\newtheorem{example}{Example}
\newtheorem{theorem}{Theorem}

\usepackage{newfloat}
\usepackage{listings}
\DeclareCaptionStyle{ruled}{labelfont=normalfont,labelsep=colon,strut=off} 
\floatstyle{ruled}
\newfloat{listing}{tb}{lst}{}
\floatname{listing}{Listing}
\title{Integrated Decision Gradients: Compute Your Attributions Where the Model Makes Its Decision}
\author {
    Chase Walker\textsuperscript{\rm 1}, 
    Sumit K. Jha\textsuperscript{\rm 2}, 
    Kenny Chen\textsuperscript{\rm 3}, 
    Rickard Ewetz\textsuperscript{\rm 1}
}
\affiliations {
    \textsuperscript{\rm 1}Department of Electrical and Computer Engineering, University of Central Florida, Orlando, FL, USA\\
    \textsuperscript{\rm 2}Knights Foundation School of Computing and Information Sciences, Florida International University, Miami, FL, USA\\
    \textsuperscript{\rm 3}Lockheed Martin Corporation, Orlando, FL, USA\\
    chase.walker@ucf.edu, jha@cs.fiu.edu, kenny.chen@lmco.com, rickard.ewetz@ucf.edu
}

\begin{document}

\maketitle

\begin{abstract}
Attribution algorithms are frequently employed to explain the decisions of neural network models. Integrated Gradients (IG) is an influential attribution method due to its strong axiomatic foundation. The algorithm is based on integrating the gradients along a path from a reference image to the input image. Unfortunately, it can be observed that gradients computed from regions where the output logit changes minimally along the path provide poor explanations for the model decision, which is called the \emph{saturation effect} problem. In this paper, we propose an attribution algorithm called integrated decision gradients (IDG). The algorithm focuses on integrating gradients from the region of the path where the model makes its decision, i.e., the portion of the path where the output logit rapidly transitions from zero to its final value. This is practically realized by scaling each gradient by the derivative of the output logit with respect to the path. The algorithm thereby provides a principled solution to the saturation problem. Additionally, we minimize the errors within the Riemann sum approximation of the path integral by utilizing non-uniform subdivisions determined by adaptive sampling. In the evaluation on ImageNet, it is demonstrated that IDG outperforms IG, Left-IG, Guided IG, and adversarial gradient integration both qualitatively and quantitatively using standard insertion and deletion metrics across three common models.
\end{abstract}

\section{Introduction}
The access to internet-scale data and compute power has fueled the success of black box neural network models for applications such as disease detection \cite{ml_disease_survey}, image synthesis \cite{synthesis}, and protein folding \cite{colabfold}. The phenomenal performance of these networks comes from the large number of parameters and non-linear interactions among them. The complex and high dimensional dynamics makes it difficult to understand and visualize why a neural network makes a particular decision. To establish trustworthiness in neural network models, noteworthy research efforts have been devoted to interpretability and explainability \cite{xai_survey}. Attribution methods provide model explanation by computing the contribution of each input feature to a model decision. Attribution methods broadly fall into perturbation based \cite{occlusion_deconv,lime}, backpropagation based \cite{guided_backprop,gradcam}, and gradient based methods \cite{Simoyan-et-al:saliency,Sundararajan:IG}. Gradient based methods are promising due to their strong axiomatic foundation, and model-agnostic implementation \cite{Sundararajan:IG}. 

Gradient based methods compute attribution maps by capturing the gradients at the model inputs with respect to the model outputs~\cite{Simoyan-et-al:saliency}. However, gradients computed with respect to important input pixels may be zero due to the non-linear activation functions. Integrated Gradients (IG) solved this problem by integrating the gradients along a path from a baseline reference image to the input image \cite{Sundararajan:IG}. Unfortunately, it can be observed that gradients from regions of the path where the output logit changes minimally (e.g. is saturated) provide poor explanations for the model decision~\cite{Miglani-et-al:LIG}. This phenomena is called the \emph{saturation effect} problem. Solution templates to solve the saturation problem include: selecting non-straight-line paths~\cite{Kapishnikov-et-al:GIG, AGI}, path truncation~\cite{Miglani-et-al:LIG}, post processing methods that use thresholding~\cite{Kapishnikov-et-al:XRAI}, averaging across blurred inputs~\cite{smoothgrad}, and redefining the model \cite{jha2021smoother, jha2022shaping}. While these methods improve attribution quality, they do not provide a principled solution to the saturation problem. 

\begin{figure*}[t]
    \centering
    \includegraphics[width=0.95\textwidth]{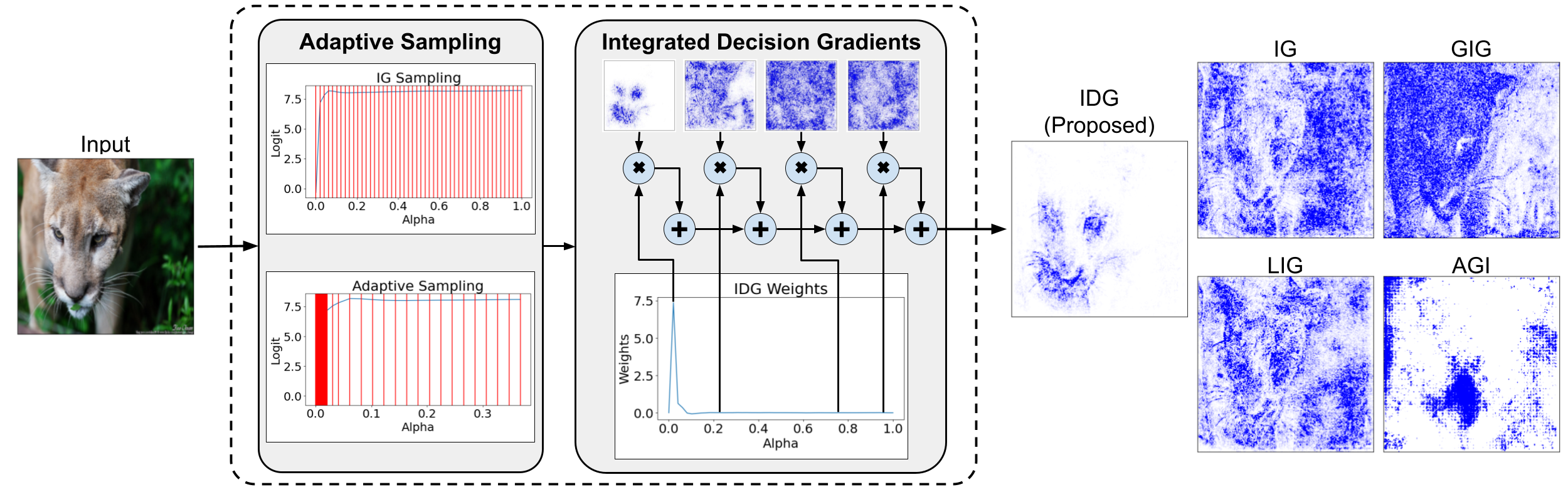}
    
    \caption{(left) An overview of the adaptive sampling algorithm, and the IDG attribution method. (right) A preliminary visual comparison of IDG with IG~\cite{Sundararajan:IG}, LIG~\cite{Miglani-et-al:LIG}, GIG~\cite{Kapishnikov-et-al:GIG}, and AGI~\cite{AGI}.}
    \label{fig:block_diagram}
\end{figure*}

In this paper, we propose a new path integral attribution method called Integrated Decision Gradients (IDG). We call the portion of the path where the output logit rapidly transitions from zero to its final value the \emph{decision region}. IDG focuses on integrating gradients from the decision region of the path integral. This is realized by scaling each gradient by the derivative of the output logit with respect to the path. The scaling factor rewards gradients in the decision region and penalizes gradients from outside the decision region. The main contributions of this paper are summarized as follows: 
\begin{itemize}
    \item We propose IDG, a new attribution method that provides a principled solution to saturation by satisfying the IG axioms and a new path integral sensitivity axiom. 
    \item We present an adaptive sampling technique to select non-uniform subdivisions for the Riemann approximation of the path integral. The non-uniform subdivisions reduce computational errors (and runtime overheads) compared with using uniform subdivisions.
    \item Compared with IG~\cite{Sundararajan:IG}, Left-IG (LIG)~\cite{Miglani-et-al:LIG}, Guided IG (GIG)~\cite{Kapishnikov-et-al:GIG}, and Adversarial Gradient Integration (AGI)~\cite{AGI}, IDG improves in both qualitative and quantitative results.
\end{itemize}%

The remainder of the paper is organized as follows: related work is examined in Section~\ref{section:related_work}, the IDG attribution method in Section~\ref{section:IDG}, the adaptive sampling algorithm is proposed in Section~\ref{section:sampling}, experimental evaluation is presented in Section~\ref{section:experiments}, and the paper is concluded in Section~\ref{section:discussion}.

\section{Related Work}
\label{section:related_work}

In this section, we first review the limitations of directly using gradients as attributions. Next, we review IG and assess the saturation effect problem within path integrals. 

\subsection{Limitations of Gradients as Attributions}
\label{section:saturation}
Attributions are defined to be the contribution of each input feature to the model output decision. An attribution method satisfies the axiom of \emph{sensitivity} if a single feature that differs between a baseline and input - which produce different output predictions - is given a non-zero attribution. Additionally, if a neural network is not affected by changing a variable, then that variable's attribution shall be zero~\cite{Sundararajan:IG}. Computing the gradient of the inputs with respect to the output logit is a promising method for computing attributions~\cite{Simoyan-et-al:saliency}. However, the use of non-linear activation functions causes the sensitivity axiom to be violated~\cite{Sundararajan:IG}, which is shown in Example 1 below.

\begin{example}
    Consider a function $F = 1 - ReLU(1 - x)$, a baseline $x' = 0$, and an input $x = 2$. For $x' = 0$, the function $F$ is equal to $0$, and for $x = 2$, the function $F$ is equal to $1$.
    Since changing $x$ from $0$ to $2$ affects the output of $F$, the attribution w.r.t. the feature $x$ should be non-zero. 
    However, $\partial F/\partial x = 0$ at $x = 2$, which results in an attribution of $0$~\cite{Sundararajan:IG}. 
    \label{grad_sensitivity}
\end{example}

Integrated gradients offers a solution to computing attributions that satisfies the sensitivity axiom. 

\subsection{Integrated Gradients}
Integrated Gradients computes attributions by integrating gradients on a straight line between a reference image and an input image~\cite{Sundararajan:IG}. Let $F$ be the function realizing the output logit of interest. $IG_i$ with input image $x$ is mathematically defined using a path-integral~\cite{Sundararajan:IG}, as follows:  
\begin{equation}
    \resizebox{0.90\linewidth}{!}{$
    \displaystyle
        IG_i(x) = (x_i - x'_i) \times \int_{\alpha = 0}^{1} \frac{\partial F(x'_i + \alpha (x_i - x'_i ))}{\partial x_i} d\alpha,
        \label{eq:IG}
    $}
\end{equation}%
where $x'$ is a black baseline image, $\alpha \in [0,1]$ parameterizes the straight-line path between $x'$ and $x$, $x_i$ and $x'_i$ represent a single pixel of their respective images, and $IG_i$ is therefore the attribution of pixel $i$ in the input image. 

\begin{figure}[!t]
    \centering
    \includegraphics[width=0.95\linewidth]{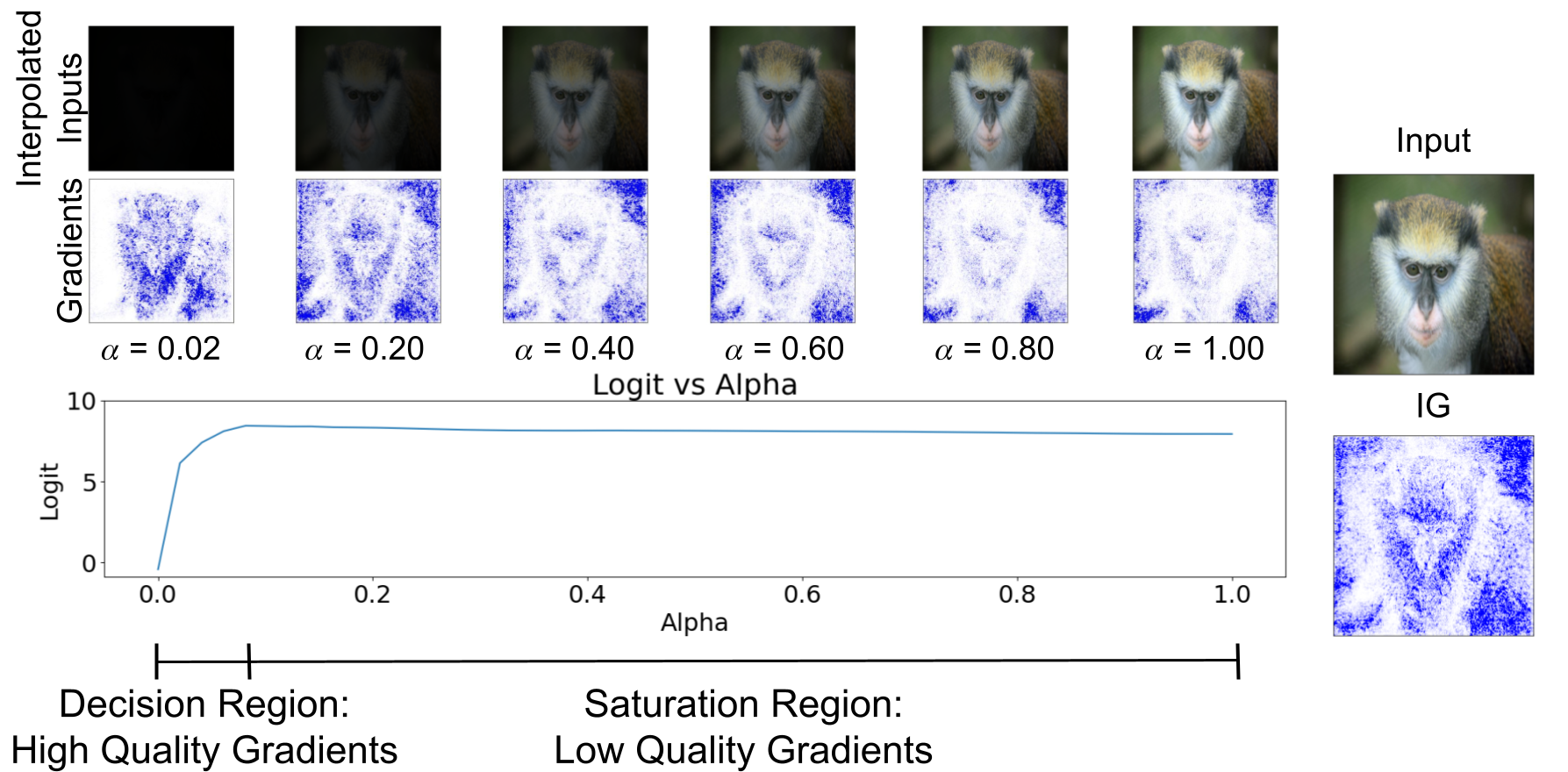}    
    \caption{The figure illustrates the IG attribution method and saturation effects within path integrals. The top row shows interpolated inputs and the second row shows the corresponding gradients. The IG attribution map (shown to the right) is the average of the gradients. 
    The third row shows the logit-$\alpha$ curve, which defines the decision and saturation regions. It can be observed that the gradients from the decision region are of higher quality than the saturation region. }
    \label{fig:ig}
\end{figure}

The IG attribution method is illustrated in Figure~\ref{fig:ig}. The top row shows interpolated inputs, the second row shows the corresponding input gradients, the third row visualizes the output logit with respect to the path. The IG attribution map is equal to the sum of the gradients in the second row. The use of a path-integral ensures that gradients from regions of $F$ where $\partial F/\partial x_i$ is non-zero are computed. In Example~\ref{grad_sensitivity}, IG will compute gradients from the region $[0,$ $1]$, where $\partial F/\partial x = 1$. The resulting attribution w.r.t. $x$ is $2$, i.e., the attribution is non-zero and sensitivity is satisfied. Nevertheless, many attributions computed using IG are still noisy due to saturation effects~\cite{Miglani-et-al:LIG}. 

\subsection{Saturation Effects within Path-Integrals}

To introduce and understand the \emph{saturation effect} problem within path-integrals, we examine the performance of the IG attribution method in Figure~\ref{fig:ig}. We study the quality of the computed gradients with respect to the decision and saturated regions of the path integral. It can be observed that (i) gradients from the saturation regions are of low quality and (ii) gradients from the decision region are of high quality. The conclusion is rather straight forward to understand.
If the model output does not increase while moving $\triangle \alpha$ along the path, it is intuitive that the corresponding gradients are not important to the model decision. Conversely, if the output logit changes rapidly while moving $\triangle \alpha$ along the path, those gradients have a strong impact on the model decision.   

This raises the rudimentary question: Is it possible to design a path integral that focuses on computing gradients from the region where the model decision is made and the highly informative gradients are located? It can, for example, be observed in Figure \ref{fig:ig} that the gradients computed at $\alpha=0.02$ alone provide an excellent explanation for the model decision. 

\section{Integrated Decision Gradients}
\label{section:IDG}
In this section, we propose a new attribution method called Integrated Decision Gradients (IDG). We outline the motivation behind the design of IDG, explain the concept of importance factors, and provide the definition as well as a visualization of IDG.  

\subsection{Motivation}
Path integrals integrate gradients from a reference image to an input target image. A fundamental challenge is to determine the ideal importance of each gradient. 
Based on the analysis in the previous section, we define a new \emph{sensitivity axiom} for path integrals. Next, we introduce the concept of an importance factor, which is used to construct an attribution algorithm that satisfies the axiom.  

\begin{figure}
    \centering
    \includegraphics[width=0.95\linewidth]{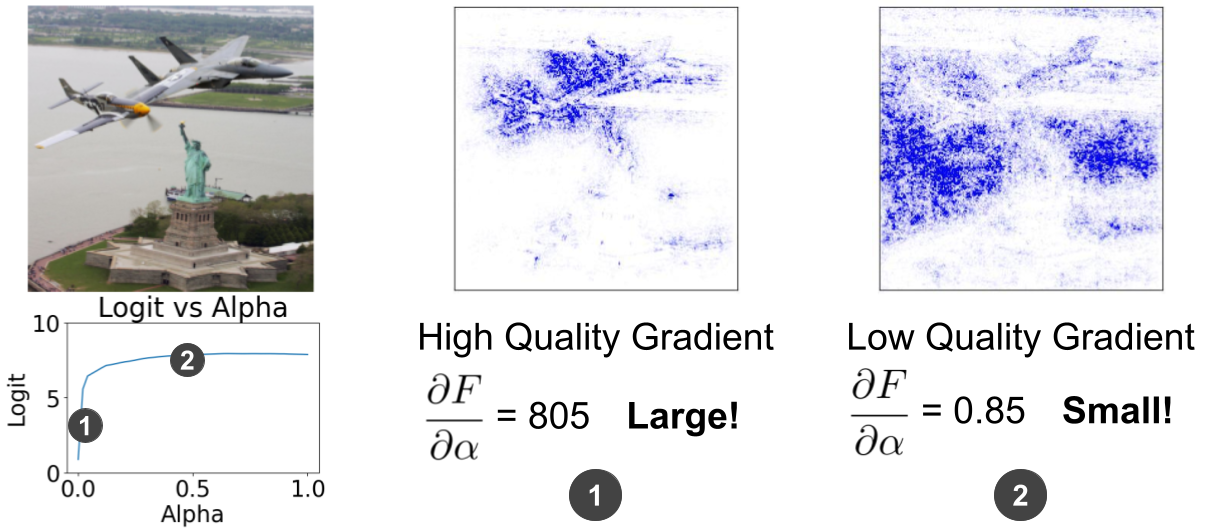}
    \caption{An illustration of the relationship between importance factor magnitude and gradient quality. Higher importance factors are directly related to higher quality gradients.}
    \label{fig:importance}
\end{figure}

\begin{figure*}[!t]
    \centering
    \includegraphics[width=0.95\textwidth]{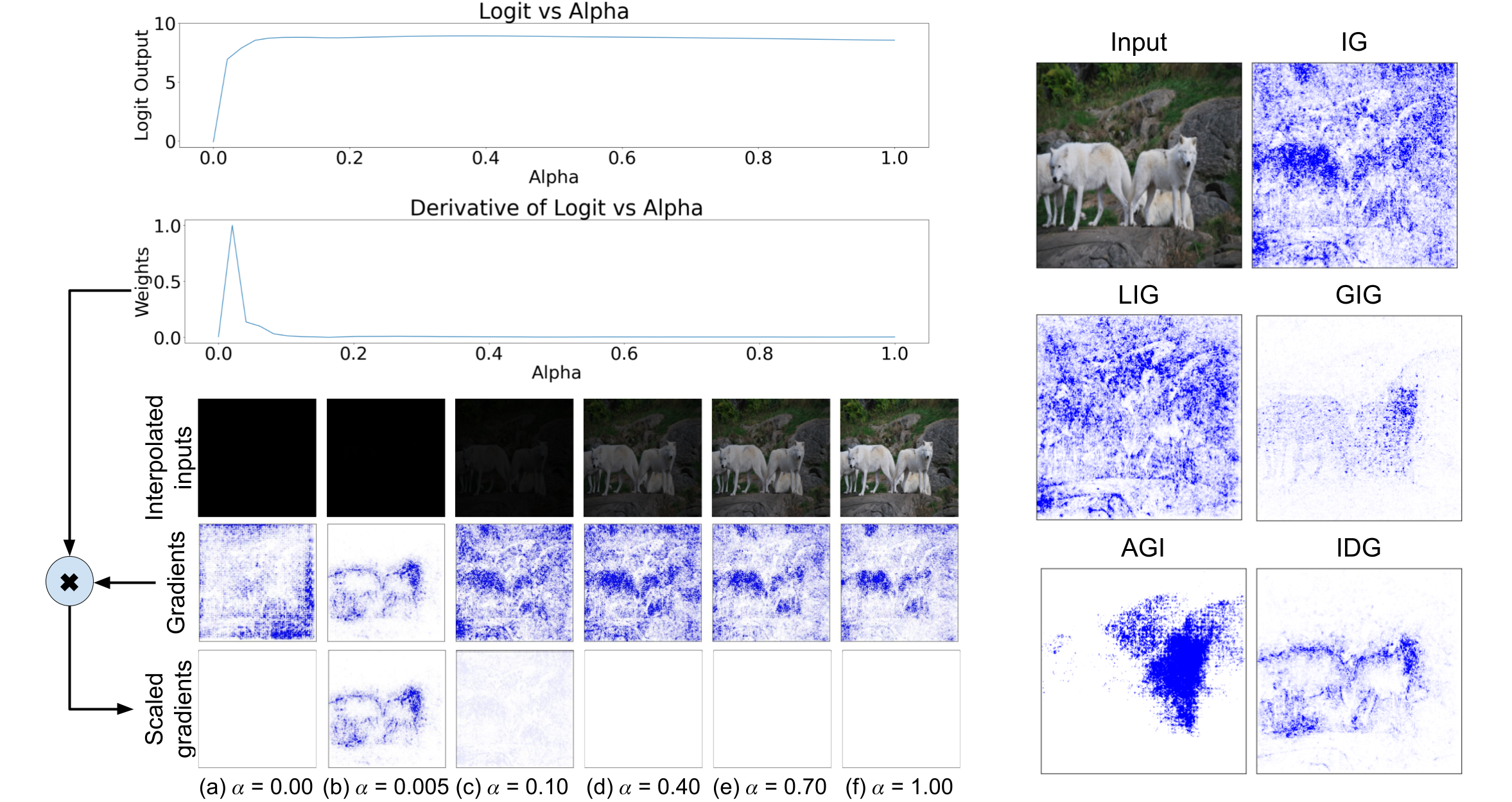}
    
    \caption{A full visualization of how IDG uses importance factors to eliminate saturation effects. The top row shows the logit-$\alpha$ curve. The next row shows the derivative of the curve, i.e., the importance factors with respect to $\alpha$. The third row shows the interpolated images, the fourth shows the associated gradients, and the bottom row shows these gradients scaled with the corresponding importance factors. The right side shows the input image, and the attributions computed using IG~\cite{Sundararajan:IG}, LIG~\cite{Miglani-et-al:LIG}, GIG~\cite{Kapishnikov-et-al:GIG}, AGI~\cite{AGI}, and IDG.}
    \label{fig:IDG_visualization}
\end{figure*}

\paragraph{Axiom: Sensitivity (path integrals)} 

Let $F$ be the output of a neural network. For every point within a path integral parameterized by $\alpha$, when $\partial F/\partial \alpha$ is equal to zero, an attribution method satisfies Sensitivity (path integrals) if there is no contribution to the attribution result. If $\partial F/\partial \alpha$ is non-zero, the contribution to the attribution result is non-zero.  

None of the existing attribution methods based on path integrals satisfy this axiom~\cite{Sundararajan:IG,Miglani-et-al:LIG,Kapishnikov-et-al:GIG,AGI}. The traditional IG method places an equal weight on all gradients~\cite{Sundararajan:IG}, even those that occur in the saturation region where $\partial F/\partial \alpha = 0$. The Left-IG attribution attempts to solve this by truncating the path integral after the output logit has reached $90\%$ of its final value~\cite{Miglani-et-al:LIG}. This assigns a weight of zero and one to gradients from the approximate saturation and decision regions respectively, which does not guarantee that the axiom is satisfied. GIG and AGI use non-straight line paths that attempt to avoid integrating gradients from saturated regions~\cite{Kapishnikov-et-al:GIG, AGI}, which does also not guarantee that the Sensitivity (path integrals) axiom is satisfied.  

To satisfy the axiom, we conjecture that the importance of each gradient should be proportional to the impact on the model output, which is conceptually shown in Figure \ref{fig:importance}. Inspired by this, we define an \emph{importance factor}, as follows: 
\begin{equation}
   IF(\alpha) = \frac{\partial F(x' + \alpha(x - x'))}{\partial \alpha},
    \label{eq:importance}
\end{equation}
where $IF(\alpha)$ is the importance of the gradient computed at $\alpha$. Next, we define an attribution method that satisfies the Sensitivity (path integrals) axiom (proof in Section \ref{section:properties}) by scaling each gradient with the importance factor in Eq~(\ref{eq:importance}).

\subsection{Definition of Integrated Decision Gradients}
In this subsection, we formally define the IDG attribution algorithm. Given a neural network represented by function $F: R^{n} \to [0,1]$, an input vector $x$, and given $F$ exists over the range $\alpha \in [0,1]$, IDG assigns an importance factor to each input feature $x_i$ with respect to the model output, using the following equation: 


\begin{equation}
    \begin{split}
        IDG_i(x) = (x_i - x'_i) & \times \underbrace{\int_{\alpha = 0}^{1} \frac{\partial F(x'_i + \alpha(x_i - x'_i ))}{\partial x_i}}_{\textrm{Traditional IG}}
        \\ & \times \underbrace{ \frac{\partial F(x'_i + \alpha(x_i - x'_i ))}{\partial \alpha}}_{\textrm{Importance Factor}} d\alpha.
    \end{split}
    \label{eq:IDG}
\end{equation}
    
The IDG attribution method is equivalent to IG in Eq~(\ref{eq:IG}) but with each gradient scaled with the importance factor in Eq~(\ref{eq:importance}). The importance factor is equivalent to the derivative of the logit-$\alpha$ curve in the bottom of Figure~\ref{fig:ig}. The importance factors scale-up high quality gradients from the decision region and scale-down low quality gradients from saturated regions, respectively.  

The integral is practically computed following IG, using the Riemann sum approximation~\cite{Sundararajan:IG}, as follows: 
\begin{equation}
    \resizebox{.89\columnwidth}{!}{$
        \displaystyle
        IDG_i(x) = \frac{(x_i - x'_i)}{m} \times \sum_{k = 1}^{m} \frac{\partial F(x'_i + \frac{k}{m} (x_i - x'_i ))}{\partial x_i} \frac{\partial F}{\partial \alpha} \ d \alpha
    $},
    \label{eq:IDG_sum}
\end{equation}%
where $m$ is the number of steps for approximation. We will further discuss the selection of the step size and its impact on the approximation error in Section~\ref{section:sampling}.  

We illustrate IDG with an example in Figure~\ref{fig:IDG_visualization}. First, looking at the left side of the figure, the top row shows the logit-$\alpha$ curve associated with the input image. The second row shows the derivative of this curve, i.e., $\partial F/\partial \alpha$ in Eq (\ref{eq:importance}). The third row shows the interpolated inputs for selected alpha values and the fourth row shows the gradients computed by IG for these inputs. The last row visualizes the effect of IDG by scaling the gradients above by the importance factors from the second graph. The importance factors scale up the magnitude of the gradients from the decision region while scaling down the magnitude of the gradients from the saturated regions. In the figure, it can be observed that, in particular, the attributions from $\alpha = 0.005$ are scaled up. On the right of the figure, we show the original image, and the attributions generated by IG, LIG, GIG, AGI, and IDG. The attributions computed using IDG are substantially less noisy than all competitors. We note that GIG has a low amount of noise, but IDG has more focused attributions. 

\subsection{Axiomatic Properties of IDG}
\label{section:properties}

In~\citeauthor{lundstrom}, it was shown that the axiomatic properties of IG such as Completeness, Sensitivity, Implementation Invariance, and Linearity only hold when assuming a monotonically increasing path and non-decreasing $F$. We show in the supplementary materials that IDG satisfies the exact same axiomatic properties under the same assumptions. Next, we turn our attention to proving that IDG satisfies Sensitivity (path integrals). 

\begin{theorem}
    IDG is the sole path method to satisfy Sensitivity (path integrals) through the use of the importance factor.
\end{theorem}

\begin{figure*}[!t]
    \centering
    \includegraphics[width=0.9\textwidth]{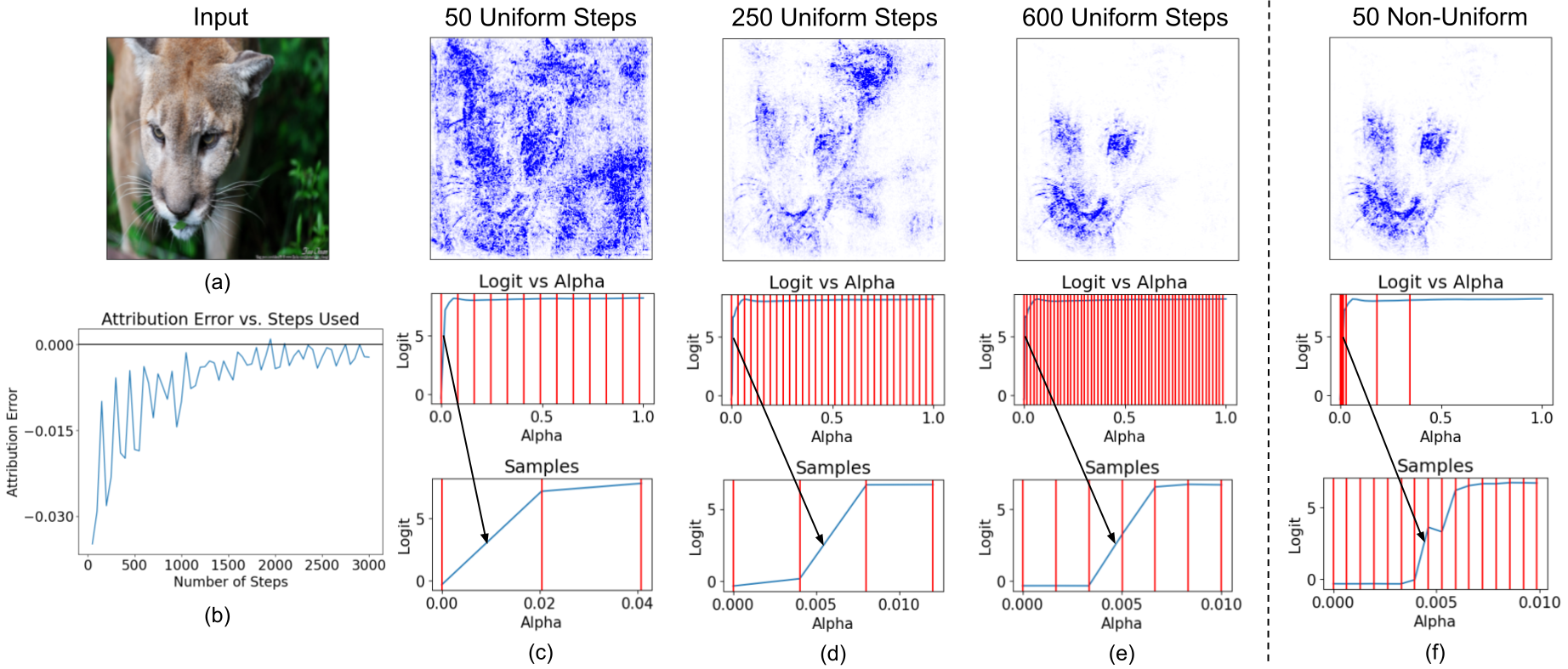}
    
    \caption{This figure shows the motivation for the adaptive sampling algorithm. The image (a) is the input to the attributions in the figure. The graph (b) demonstrates how the attribution error decreases as step count increases. Columns (c), (d), and (e) of attributions and graphs show the relationship between sample locations and IDG quality as $50$, $250$, and $600$ steps are used respectively. We show that as the number of steps increases, the quality of IDG grows greatly, influencing the adaptive sampling algorithm. Lastly, column (f) shows the equivalent result of column (e) achieved by using adaptive sampling with 50 steps.}
    \label{fig:IDG_steps}
\end{figure*}

\begin{proof}
    Consider the neural network $F$ which is continuous and differentiable over $\alpha$ such that $\partial F / \partial \alpha$ is defined. By definition, an IDG attribution at $\alpha$ along the path is $IDG(\alpha) = \partial F / \partial x \times \partial F /\partial \alpha$. $(x-x')$ is ignored as it is a post-processing factor applied to the attribution which is affected by proper baseline selection. 

    When $\partial F / \partial \alpha = 0$, the $IDG(\alpha)$ attribution is clearly zero. When $\partial F /\partial \alpha \neq 0$, then $\partial F/\partial x \neq 0$ for at least one feature and the resulting $IDG(\alpha)$ attribution will be non-zero. Therefore it follows, by definition, that IDG satisfies Sensitivity (path integrals).
\end{proof}

\section{Adaptive Sampling Algorithm}
\label{section:sampling}
In this section, we first analyze the errors within the Riemann sum approximation of the IDG path integral for uniform subdivisions. Next, we propose an adaptive sampling technique to minimize the approximation errors using non-uniform subdivisions. In the supplementary materials, we show that the adaptive sampling only creates improvements with IDG and its impact on regular IG is minor.

\subsection{Motivation}

The errors within the Riemann approximation of the IDG path integral can be calculated, as follows: 
\begin{equation}
    \epsilon(x,n) = \lim_{m\to\infty} IDG_i(x,m) - IDG_i(x,n),
    \label{eq:int_app}
\end{equation}
where $\epsilon(x_i,n)$ is the approximation error for attribution $x_i$ when computing the integral with $n$ uniform subdivisions. $n$ and $m$ are the number of steps used within the Riemann sum approximation in Eq~(\ref{eq:IDG_sum}). 

We analyze the approximation error and the impact on the attributions in Figure~\ref{fig:IDG_steps}. The graph (b) shows the average error across all the pixels in the attribution map with respect to the number of used steps $n$. Since a low step count results in a lack of samples in the decision region, a large number of steps are required for a good approximation. The image (a) is the input for the four columns (c), (d), (e), and (f) of attributions and graphs. The columns show the quality of the attributions with respect to the number of steps and type of subdivision. It is observed from the graphs that taking more samples in the decision region greatly improves  IDG attribution quality. Therefore, to obtain high IDG quality without a prohibitive number of steps, we design a new adaptive sampling algorithm - seen in Figure~\ref{fig:IDG_steps} (f) - that uses non-uniform subdivisions concentrated on the decision region.

\subsection{Adaptive Sampling Methodology}
\label{section:algorithm_def}

It is desirable to sample the high quality gradients that lie in the decision region to improve the quality of the attained attributions. In Algorithm \ref{alg:IDG_AS}, we show how the adaptive sampling algorithm is used with IDG. Our approach is based on first pre-characterizing the logit-$\alpha$ curve with $N$ uniform subdivisions in lines 3 - 7. Next, $M$ subdivisions are non-uniformly distributed within the $N$ regions based on logit growth and IDG is calculated 
in lines 8 - 15. Because there are $M$ total samples, line 11 executes $O(N + M)$ times. In practice it is best if $N = M$ (this is shown in the supplementary materials) therefore the algorithm runtime is $O(N)$.

As seen in Figure~\ref{fig:IDG_steps} (e) and (f), combining this adaptive sampling algorithm with IDG creates attributions as strong as IDG with 600 steps while only using 50 steps. Figure~\ref{fig:block_diagram} provides a high-level overview of this new IDG process. The figure shows that when given an input image and a number of steps, the adaptive sampling algorithm calculates non-uniform subdivisions based on logit growth. These are then used as input for IDG where the gradient at each location is calculated and then weighted, producing the final attribution. In this figure, the IDG sampling graph shows that $31$ out of $50$ samples are placed in the decision region $\alpha \in [0.0, 0.2]$, where the logit changes from $0$ to $7.2$. 

\renewcommand{\algorithmiccomment}[1]{#1}
\begin{algorithm}[!t]
    \textbf{Input:} Model $F$, image $x$, baseline $x'$, pre-characterization steps $N$, number of IDG steps $M$\\
    \textbf{Output:} An attribution map $A$
    \caption{Computing IDG with Adaptive Sampling} 
    \label{alg:IDG_AS}
    \begin{algorithmic}[1]
        \STATE $x^0 = x'$
        \STATE $x^{N-1} = x$
        \STATE $samples[0] = 0$ \\
        \COMMENT {// Pre-characterization of logit-$\alpha$ curve}\\
        \FOR{$i = 0$ to $N - 1$ }
            \STATE $x^{i + 1} = x' + \frac{i}{N} \times (x - x')$
            \STATE $samples[i + 1] = round(\frac{F(x^{i + 1}) - F(x^i)}{F[x^{N-1}]- F[x^0]} \times M$)
        \ENDFOR 
        \\
        \COMMENT{// Computation of IDG with non-uniform samples}\\
        \FOR{$i = 0$ to $N$} 
            \FOR{$j = 0$ to $samples[i]$}
                \STATE $\alpha = \frac{i}{N} + \frac{j}{N \times samples[i]}$
                \STATE $x^i = x' + \alpha \times (x - x')$
                \STATE $IDG[i] = \frac{\partial F(x^i)}{\partial x} \times \frac{\partial F(x^i)}{\partial \alpha} \times \frac{1}{N \times samples[i]}$ 
            \ENDFOR
        \ENDFOR

        \RETURN $A = mean(IDG)$
    \end{algorithmic}
\end{algorithm}

\section{Experimental Results}
\label{section:experiments}
In this section, we will evaluate the effectiveness of the proposed method. We perform our experiments in PyTorch using the 2012 validation set of ImageNet~\cite{imagenet} on NVIDIA A40 GPUs.
The attributions computed using Algorithm \ref{alg:IDG_AS} are called IDG. We compare our method with IG \cite{Sundararajan:IG}, LIG \cite{Miglani-et-al:LIG}, GIG \cite{Kapishnikov-et-al:GIG}, and AGI \cite{AGI}. We use Captum for the implementation of IG, whereas LIG, GIG, and AGI are taken from their respective repositories~\cite{captum, LIG-code, GIG-code, AGI-code}. We evaluate the quality of the computed attributions both quantitatively and qualitatively.

In Table \ref{tab:results}, we quantitatively evaluate the attributions using standard perturbation testing which measures the importance of the pixels in an attribution via an area under the curve (AUC) score. Four tests are used with three insertion methods and one deletion method from the authors of RISE and XRAI~\cite{petsiuk:RISE, Kapishnikov-et-al:XRAI} which are described in Section \ref{section:tests}. The table compares the computed attribution quality for the first $5000$ images of the ImageNet dataset such that five images are taken from each of the $1000$ classes. The five attribution methods are evaluated with three models trained on ImageNet. We selected ResNet101 (R101), ResNet152 (R152), and ResNeXt (RNXT) as pre-trained models from PyTorch and use the newest ImageNet weights available (V2 for the ResNet models and V1 for ResNeXt)~\cite{resnet101, RNXT, pytorch}.

Qualitatively, we present five selected examples in Figure \ref{fig:IG_comparisons} for the the ResNet101 model generated with the method parameters explained below. We provide a larger, random selection of examples in the supplementary materials.

Inputs are reshaped to ($224$, $224$) for all three presented models. This image processing follows the attribution documentation provided by Captum~\cite{captum}. The RISE, AIC, and SIC tests use the default parameters found from their respective repositories~\cite{RISE-code, AIC-code}. The IG and LIG attribution methods use $50$ steps and a black baseline image. GIG uses the default parameters found at~\cite{GIG-code}. AGI uses the default parameters found at~\cite{AGI-code}. Lastly, IDG is used with 50 steps and a black baseline image. For all the methods, we use a single baseline only.

\subsection{Quantitative Evaluation Metrics}
\label{section:tests}

The evaluation metrics are built upon the intuition that the highest attribution values should correspond to those features that contribute more to the classification of the target class~\cite{petsiuk:RISE, Kapishnikov-et-al:XRAI}. The process starts from the most important pixels and starts deleting (inserting) them from the original image (to a blurred image for insertion) until only a black (the original) image remains. At each step, the softmax score (or accuracy) is calculated. This gives us an ROC curve from base image to final image, which is used to compute the AUC score for a given attribution. This AUC value is computed for each image and then averaged out over the entire test data selection. For the insertion game, a higher AUC score indicates a better attribution and for the deletion game, a lower AUC score indicates better performance. The two sets of methods presented from Petsiuk, et al and Kapishnikov, et al. take different approaches to the insertion process~\cite{petsiuk:RISE, Kapishnikov-et-al:XRAI}. 

In RISE, the insertion (deletion) test which starts (ends) with a Gaussian blurred (black) image~\cite{petsiuk:RISE}. In their implementation, pixels are added (deleted) in equal amounts during the test process. Given an NxN image, the test will change the image by N pixels at a time over N steps. 

\begin{table*}[!t]
  \centering
  \begin{tabular}{llrrrrr}
    \toprule
    Metric & Model & \multicolumn{1}{l}{IG \shortcite{Sundararajan:IG}} & \multicolumn{1}{l}{LIG \shortcite{Miglani-et-al:LIG}} & \multicolumn{1}{l}{GIG \shortcite{Kapishnikov-et-al:GIG}} & \multicolumn{1}{l}{AGI \shortcite{AGI}} & \multicolumn{1}{l}{IDG (ours)} \\
    \midrule
                        & R101 & 0.571 & 0.589 & 0.626 & 0.675 & \textbf{0.701} \\
    AIC ($\uparrow$)    & R152 & 0.575 & 0.616 & 0.646 & 0.686 & \textbf{0.718} \\
                        & RNXT & 0.580 & 0.611 & 0.634 & 0.654 & \textbf{0.730} \\
    \midrule
                        & R101 & 0.498 & 0.522 & 0.559 & 0.609 & \textbf{0.638} \\
    SIC ($\uparrow$)    & R152 & 0.508 & 0.552 & 0.582 & 0.619 & \textbf{0.659} \\
                        & RNXT & 0.478 & 0.518 & 0.532 & 0.554 & \textbf{0.620} \\
    \midrule
                        & R101 & 0.498 & 0.535 & 0.547 & 0.561 & \textbf{0.592} \\
    Insertion ($\uparrow$)   & R152 & 0.517 & 0.562 & 0.565 & 0.577 & \textbf{0.615} \\
                        & RNXT  & 0.276 & 0.299 & 0.296 & 0.307 & \textbf{0.324} \\
    \midrule
                        & R101 & 0.181 & 0.148 & 0.155 & 0.172 & \textbf{0.108} \\ 
    Deletion ($\downarrow$) & R152 & 0.202 & 0.148 & 0.164 & 0.190 & \textbf{0.118} \\
                        & RNXT  & 0.101 & 0.078 & 0.082 & 0.104 & \textbf{0.068} \\
    \bottomrule
\end{tabular}

  \caption{Comparison of path-based attributions using the AIC, SIC, insertion, and deletion tests}
  \label{tab:results}
\end{table*}

Kapishnikov, et al. present the Accuracy Information Curve (AIC) and Softmax Information Curve (SIC) in their XRAI paper~\cite{Kapishnikov-et-al:XRAI}. The AIC test gives each perturbation step a score of $0$ or $1$ for an incorrect or correct classification and SIC uses softmax as previously discussed. For pixel perturbation, these methods use a schedule that non-linearly removes groups of pixels from the image in increasingly large amounts. The last difference from the RISE insertion test is the blurring method, where the initial image is now blurred in segments, each having its own noise distribution.

\subsection{Comparison With Previous Work}

In Table \ref{tab:results}, attribution quality is evaluated using the AIC and SIC insertion metrics and the RISE insertion and deletion metrics. We use an arrow to denote if larger (arrow up) or smaller (arrow down) scores are better. The best score for each model and test type is in bold. Additionally we provide how many times a given method outperforms all other methods in the last row of the table.

It can be observed in Table \ref{tab:results} that IDG achieves a consistent improvement over IG, LIG, GIG, and AGI across all twelve of the tests presented. Comparing IDG to IG and LIG clearly indicates the ability of IDG to mitigate saturation effects in path-based methods while retaining the most important gradient information. When compared to AGI and GIG, the large margin of improvement in the scores shows that IDG presents a more complete solution to the saturation problem than these methods. Overall, IDG outperforms all of the path-based attribution methods in the comparison, achieving new state-of-the-art performance. 

\begin{figure}[ht]
    \centering
    \includegraphics[width=0.95\linewidth]{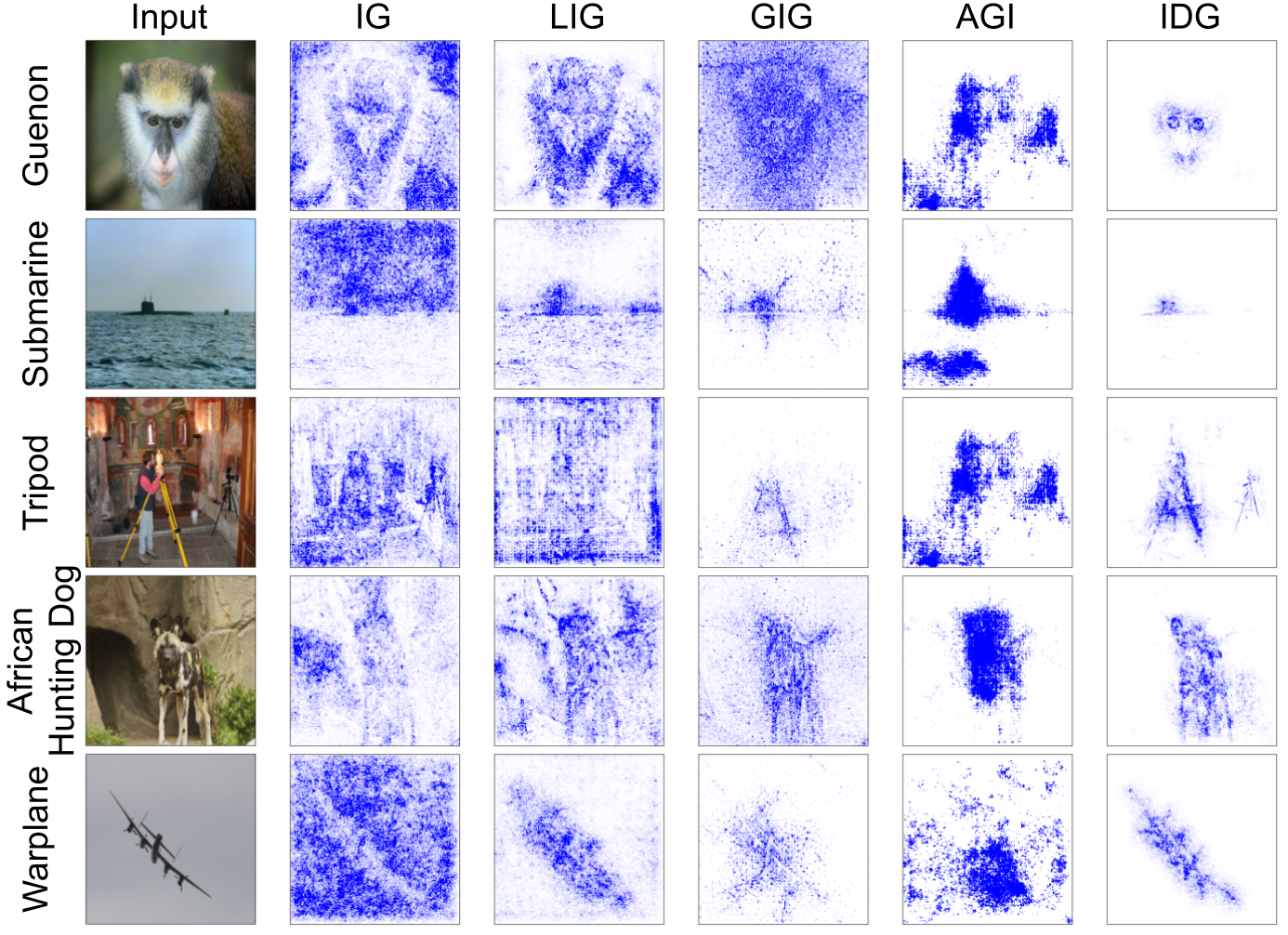}    
    \caption{Qualitative comparison of attributions computed using the IG~\cite{Sundararajan:IG}, LIG~\cite{Miglani-et-al:LIG}, GIG~\cite{Kapishnikov-et-al:GIG}, and AGI~\cite{AGI}, and IDG methods. It is seen in the selected examples that IDG solves the saturation problem and outperforms the state-of-the-art path-based attribution methods in visual quality.}
    \label{fig:IG_comparisons}
\end{figure}

For qualitative analysis, we compare IG, LIG, GIG, AGI, and IDG in Figure~\ref{fig:IG_comparisons}. All attributions are computed as previously described. The comparison is performed using images of a ``Guenon", ``Submarine", ``Tripod", ``African Hunting Dog", and ``Warplane" taken from ImageNet~\cite{imagenet}. 

Across the five selections, IDG clearly produces attributions with less noise than IG and LIG, further verifying that it solves the saturation problem present in these methods. When compared to GIG, IDG also has superior performance in all of the images. For the Tripod example, even though GIG has relatively low noise, IDG has stronger attributions on the tripod in the foreground and the one in the background as well. Lastly, when comparing to AGI, it can be seen AGI generally has low extraneous noise in the attributions. However, IDG provides tighter, and lower noise attributions on the class subject in the images, therefore the results are better.

The images clearly show that IDG improves visual quality over the other path-based methods. IDG generates attributions with less random noise, showing its ability to solve the saturation problem. Additionally, it shows its ability to outperform the methods which use non-straight-line paths. We provide an additional 50 visual comparisons in the supplementary materials. 

\section{Discussion}
\label{section:discussion}

In this paper, we propose a new attribution method called Integrated Decision Gradients (IDG). The key idea of IDG is to perform the path integral while weighting sampled gradients by their associated logit growth. This amplifies gradients from the decision region, and negates those from the saturation region, solving the saturation issue, and satisfying the Sensitivity (path integrals) axiom. In contrast, traditional IG integrates gradients between the same images while giving all gradients equal weight, saturated or not, causing the majority of saturated gradients to dominate the output. Additionally, we provide evidence that the decision region of the path integral is where the best gradients lie. With this, we present an adaptive sampling algorithm which densely samples the decision region without runtime penalty, improving IDG performance. We show qualitatively and quantitatively that IDG reaches state-of-the-art performance in the path-based attribution field. In our future work, we plan to apply IDG concepts to other attribution methods to further enhance attribution quality. We also plan to employ IDG within practical real-world applications. Our code is publicly available via \url{https://github.com/chasewalker26/Integrated-Decision-Gradients}.

\section*{Acknowledgements}
The authors were in part supported by Lockheed Martin Corp., the Florida High Tech Corridor, DARPA Co-operative Agreements \#HR00112020002, \#HR00112420004, and \#FA8750-23-2-0501, and DOE grants \#DE-SC0023494 and \#DE-SC0024576. The views, opinions and/or findings expressed are those of the authors and should not be interpreted as representing the official views or policies of those providing support for this work.

\appendix
\section{Appendix}

In this appendix we provide additional information that did not fit in the bounds of the paper. In Appendix \ref{section:IDG_path_proof} we provide the proof of the claim that IDG is a path integral from Section 3.3, guaranteeing it has the same axiomatic properties as IG. In Appendix \ref{section:AS_NM} we provide detailed analysis of the selection of $N$ and $M$ for the adaptive sampling algorithm. In Appendix \ref{section:AS} we provide further explanation of the impact of the AS algorithm, showing that IDG provides the true solution to the saturation problem. Lastly, in Section \ref{section:visual} we present 50 additional qualitative visual comparisons of our proposed method against those presented in the manuscript.

\subsection{Proof That IDG Is a Path Method}
\label{section:IDG_path_proof}

It was shown by \citeauthor{Sundararajan:IG} that all path integrals satisfy the IG axioms of Completeness, Sensitivity, Implementation Invariance, and Linearity. It was later found by \citeauthor{lundstrom} that the axioms of IG only hold for path integrals which have a monotonically increasing path and when $F$ is non-decreasing. We aim to show that IDG -- under these same assumptions -- can be defined as a path integral and has the axiomatic properties of path integrals.

\begin{proof} 
    Given a neural network $F$ which is strictly increasing, an input vector $x$, baseline $x'$, and $F$ exists over the range $\alpha \in [0,1]$, IDG is defined as follows:
    \begin{equation}
        \resizebox{0.89\linewidth}{!}{$
        \displaystyle
            IDG_i(x) = (x_i - x'_i) \times \int_{\alpha = 0}^{1} \frac{\partial F(x'_i + \alpha \times (x_i - x'_i ))}{\partial x_i} \times \frac{\partial F}{\partial \alpha} d\alpha
        $}.
        \label{eq:IDG_proof}
    \end{equation}%
    
    First, we define $\beta = F(x' + \alpha \times (x - x'))$. We propose to change the axis of integration for IDG from $\alpha$ to $\beta$ using a change of variables, i.e. converting the horizontal integration along $\alpha$ to a vertical integration along $\beta$. As $F$ is strictly increasing, its inverse $F^{-1}(x' + \beta \times (x - x' ))$ exists.

    We now define the new function $\alpha = u(\beta)$, $\beta \in [u^{-1}(0), \ u^{-1}(1)]$ and differentiate to obtain $d\alpha / d\beta = u'(\beta)$. Substitution into Eq. (\ref{eq:IDG_proof}) gives:
    \begin{equation}
        \resizebox{0.89\columnwidth}{!}{$
            \displaystyle
            IDG_i(x) = (x_i - x'_i) \times \int_{\beta = u^{-1}(0)}^{u^{-1}(1)} \frac{\partial F^{-1}} {\partial x_i} \times \frac{\partial F}{\partial \alpha} \times u'(\beta) d\beta
        $}.
        \label{eq:IDG_deriv_1}
    \end{equation}
    
    Noting that $\partial\beta = \partial F(x' + \alpha \times (x - x'))$, the following holds:
    \begin{equation}
        u'(\beta) = \left(\frac{\partial F(x' + \alpha \times (x - x' ))}{\partial\alpha}\right)^{-1}
        \label{eq:IDG_deriv_2}
    \end{equation}
    and substitution into Eq. (\ref{eq:IDG_deriv_1}) yields:
    \begin{equation}
        \resizebox{0.89\columnwidth}{!}{$
            \displaystyle
            IDG_i(x) = (x_i - x'_i) \times \int_{\beta = u^{-1}(0)}^{u^{-1}(1)} \frac{\partial F^{-1}(x'_i + \beta \times (x_i - x'_i ))}{\partial x_i} d\beta\
        $}.
        \label{eq:IDG_beta}
    \end{equation}
    This new vertical integration over $\beta$ is clearly a path integral in the same form as IG. This proves that IDG is a path integral when $F$ is strictly increasing.

    However, $F$ (as stated by \citeauthor{lundstrom}) is a non-decreasing function. We will now view $F$ as a function composed of strictly increasing and constant sub-functions. Following Sensitivity (path integrals), all constant sub-functions of $F$ will have $\partial F / \partial\alpha = 0$ and they will have zero contribution to the final attribution. We therefore can define a function $G$ which is a composition of the strictly increasing sub-functions of any function $F$, thus showing IDG is a path method for non-decreasing $F$. This guarantees that IDG has the same axiomatic properties that \citeauthor{Sundararajan:IG} state all path methods meet.
\end{proof}

\subsection{Ablation Study for Adaptive Sampling}
\label{section:AS_NM}

The adaptive sampling algorithm has two parameters $N$ and $M$. $N$ is the number of samples used in the pre-characterization of the logit-$\alpha$ curve. $M$ is the number of samples used in the computation of IDG using non-uniform subdivisions. Three types of selections of $N$ and $M$ are possible: $N < M$, $N == M$, and $N > M$. Assuming $M$ or $N$ is set to $50$, which is a common step count for path-based methods, we provide analysis of which selection provides the best result via an ablation study. 

\begin{figure}[!ht]
    \centering
    \includegraphics[width=0.99\columnwidth]{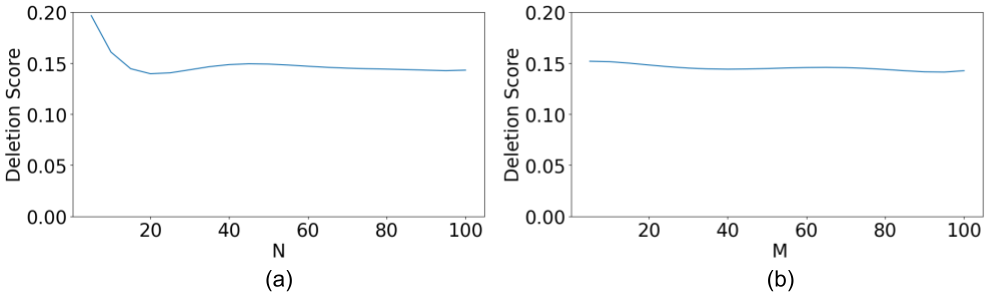}
    \caption{The change in the deletion score of IDG with AS averaged over $10$ images by varying (a) $N$ and (b) $M$. In the graphs, $N$ ($M$) is varied from $5$ to $100$ while $M$ ($N$) is set to $50$. It is seen that the most stable scores are located where $N == M$.}
    \label{fig:AS_ablation}
\end{figure}

In Figure \ref{fig:AS_ablation}, we present an ablation study on the selection of $N$ and $M$. In (a), $M$ is set to $50$ and we take the average of the deletion score \cite{petsiuk:RISE} over $10$ images as $N$ is varied from $5$ to $100$ by increments of $5$. In (b), $N$ is set to $50$ and the deletion scores are gathered as before where $M$ is varied instead. We see from graph (a) that low values of $N$ produce poor results and the transition from $5$ to $20$ results in a large drop in deletion score. We see a similar case in (b) where the score improves as $M$ increases. We note that stable performance is seen on both graphs where $N == M$.

\begin{figure}[!b]
    \centering
    \includegraphics[width=0.98\columnwidth]{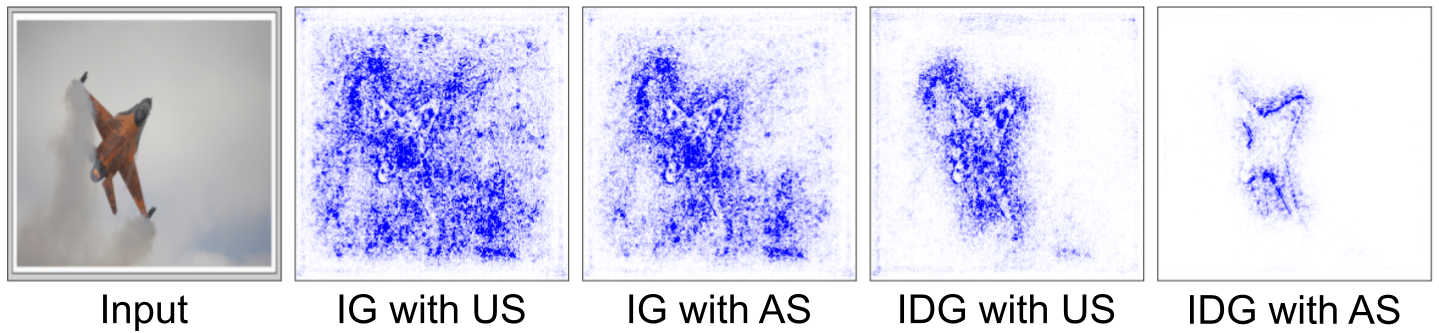}
    
    \caption{Given the input, attributions created by IG using uniform sampling (US) and AS are compared to attributions created by IDG with US and AS. Since AS applied to IG does not meaningfully improve performance over IG with US, and IDG with US provides a higher quality attribution, we determine that IDG, not AS, is the solution to the saturation problem. This is further exemplified by the improvement seen in IDG with AS, reinforcing the idea that AS gives IDG access to better gradients, but IDG is the solution to the saturation problem.}
    \label{fig:IG_AS}
\end{figure}

We conclude that selecting $N == M$ results in proper estimation of the importance factors given the $M$ IDG steps available for placement. While selecting $N < M$ may provide equally strong results, it may provide poor results without meaningful runtime improvement, therefore $N == M$ is chosen. Additionally, we note selecting an $N > M$ does not improve the score enough for the associated runtime penalty.

\subsection{IDG Is the Solution to the Saturation Problem}
\label{section:AS}

In the manuscript we present adaptive sampling as a method to improve the ability of the proposed IDG method to solve the saturation problem. Adaptive sampling takes advantage of the importance factors to perform non-uniform sampling that focuses on the region of growth. This provides a large ratio of high quality gradients to saturated gradients from which IDG can generate an attribution. However, AS alone is not a solution to the saturation problem, which we will demonstrate by evaluating IG with adaptive sampling.

In Figure \ref{fig:IG_AS}, for the given input image we compare the attributions generated by IG with uniform sampling (US), IG with adaptive sampling, IDG with uniform sampling, and IDG with adaptive sampling. When comparing IG with US and IG with AS, we see a small reduction in noise in the AS attribution, as there are inherently less saturated gradients captured when AS is applied to IG. However, due to IG equally weighting all gradients, the saturated gradients still dominate the output, illustrating that AS alone cannot solve the saturation problem. However, when viewing IDG with US compared to both IG attributions, we see a vast improvement to attribution quality, illustrating IDG's ability to solve the saturation problem. Furthermore, when AS is is applied to IDG, its ability to solve the saturation problem increases.

We quantitatively evaluate the impact of adaptive sampling (AS) versus uniform sampling (US) to back the claim that IDG is the solution to the saturation problem. In Table \ref{tab:compAS} we compare IG and IDG with US and with AS on ImageNet using ResNet 101. As IG AS does not beat IDG US, we see that AS boosts performance, but the importance factor (that defines IDG) solves the saturation problem. 

\begin{table}[!ht]
    \centering
    \resizebox{0.80\columnwidth}{!}{%
        \begin{tabular}{lrrrr}
    \toprule
    Metric & \multicolumn{1}{l}{IG US} & \multicolumn{1}{l}{IG AS} & \multicolumn{1}{l}{IDG US} & \multicolumn{1}{l}{\textbf{IDG AS}} \\
    \midrule
    Ins ($\uparrow$) & 0.513 & 0.532 & 0.546 & \textbf{0.586} \\      
    \midrule
    Del ($\downarrow$) & 0.167 & 0.146 & 0.135 & \textbf{0.110} \\
    \bottomrule
\end{tabular}

    }
    \caption{Adaptive sampling ablation study of IDG}
    \label{tab:compAS}
\end{table}

As adaptive sampling does not meaningfully improve IG performance and IDG with US provides much stronger attributions than IG, we verify that integrated decision gradients is the solution to the saturation problem. We reiterate that adaptive sampling is used to provide IDG access to gradients of a higher quality than US does, therefore improving its performance, but not acting as the solution to the saturation problem. 

\subsection{Additional Visual Comparisons}
\label{section:visual}

To validate the quantitative performance presented in the paper, we visually compare IG~\cite{Sundararajan:IG}, LIG~\cite{Miglani-et-al:LIG}, GIG~\cite{Kapishnikov-et-al:GIG}, AGI~\cite{AGI}, and IDG with a larger number of examples. We present 50 example images on pages 10 to 14. There are six columns per example. From left to right the columns are: the input image, IG, LIG, GIG, AGI, and IDG. These labels are provided above the columns on each page and the class of the input image is provided to its left. The images are from the  ImageNet validation set. We use ResNet101 on pages 10 - 12, ResNet152 on page 13, and ResNeXt on page 14 \cite{RNXT}. The attributions are generated with the same parameters as the quantitative testing.

The presented attributions are analyzed visually. A stronger attribution is defined by reduction of noise in areas irrelevant to the object of the image, and stronger attribution (darker color) in areas where the object exists. After visual analysis, we believe IDG presents better attributions than all of the methods presented for a majority of the provided examples. This thorough qualitative analysis provides further proof of the strength of the proposed IDG method.

\clearpage

\includepdf[pages=1-3, pagecommand={}]{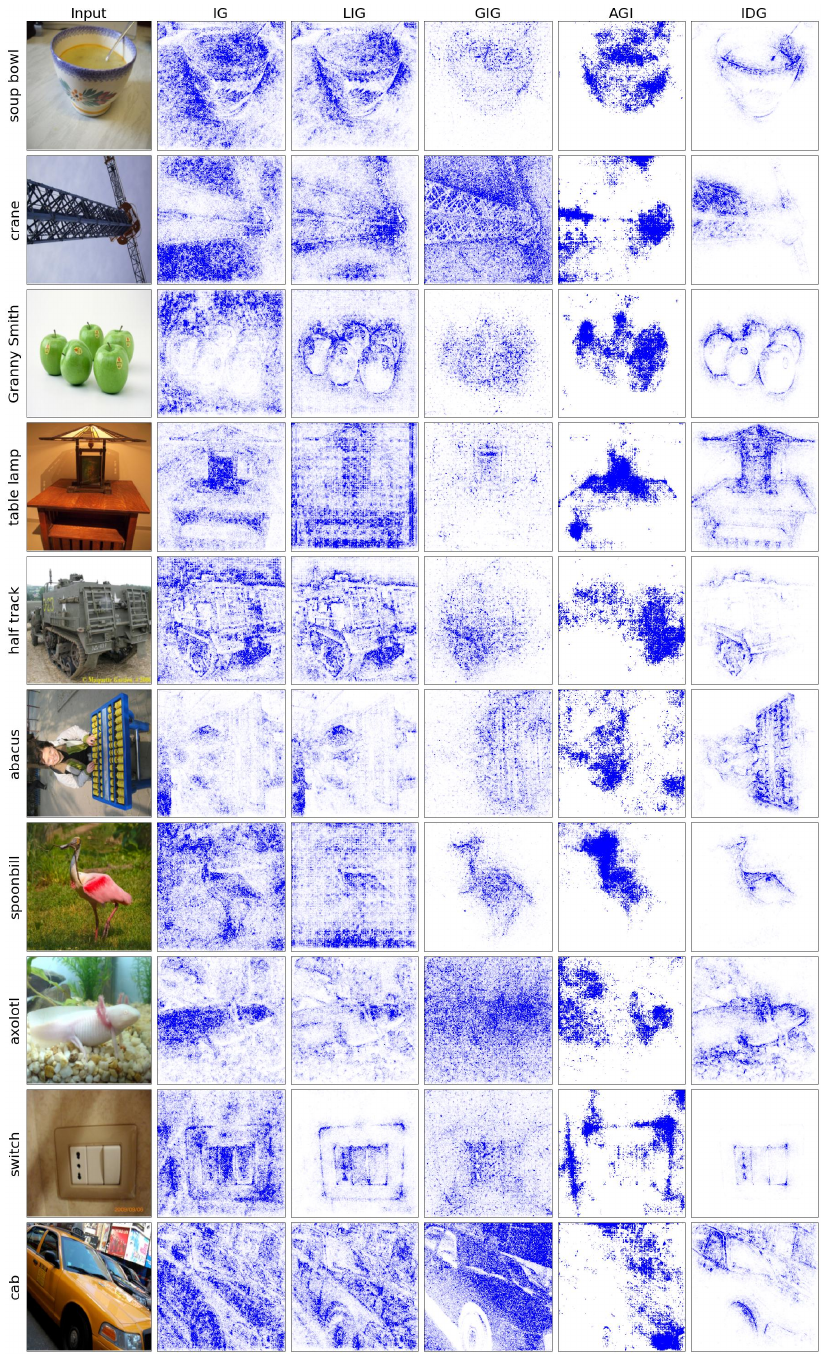}
\includepdf[pages=-, pagecommand={}]{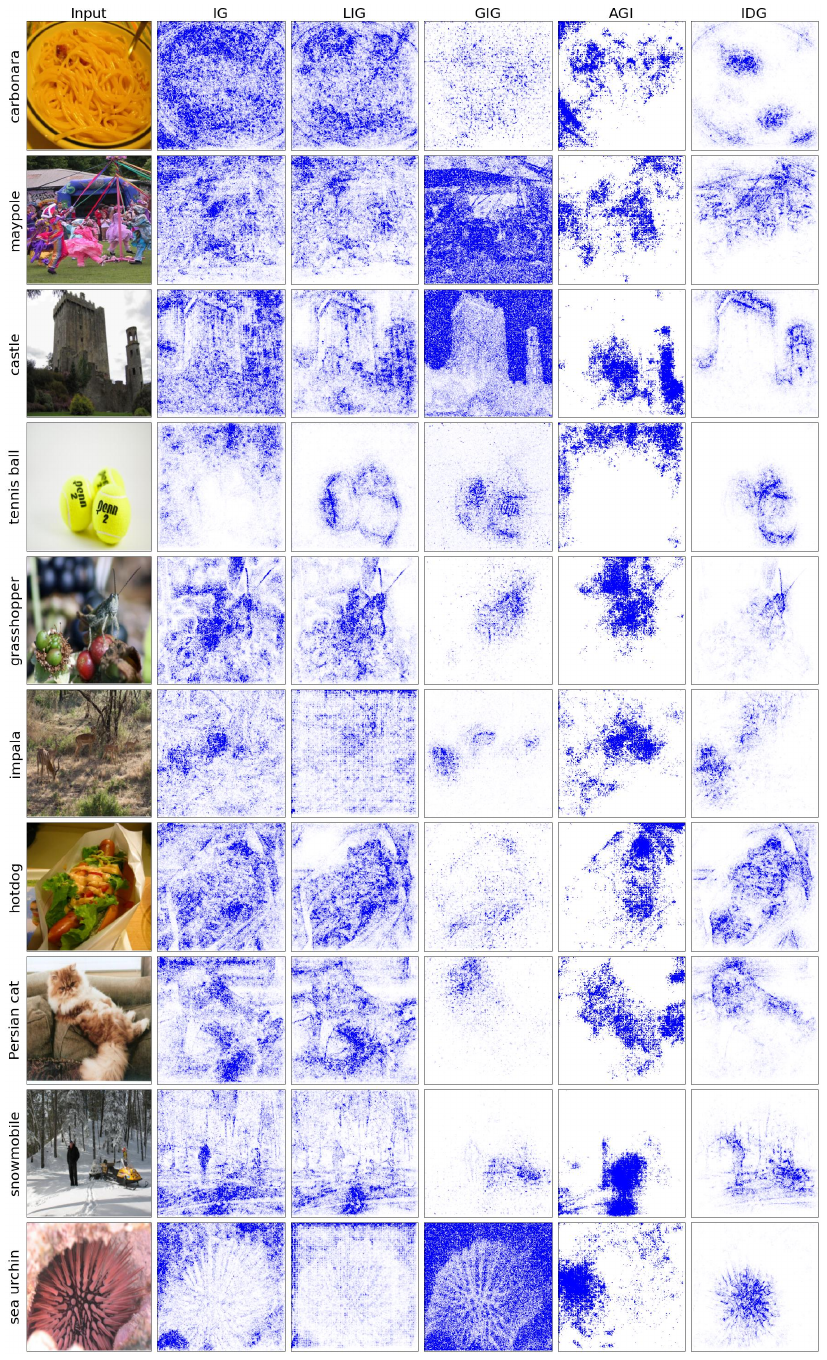}
\includepdf[pages=-, pagecommand={}]{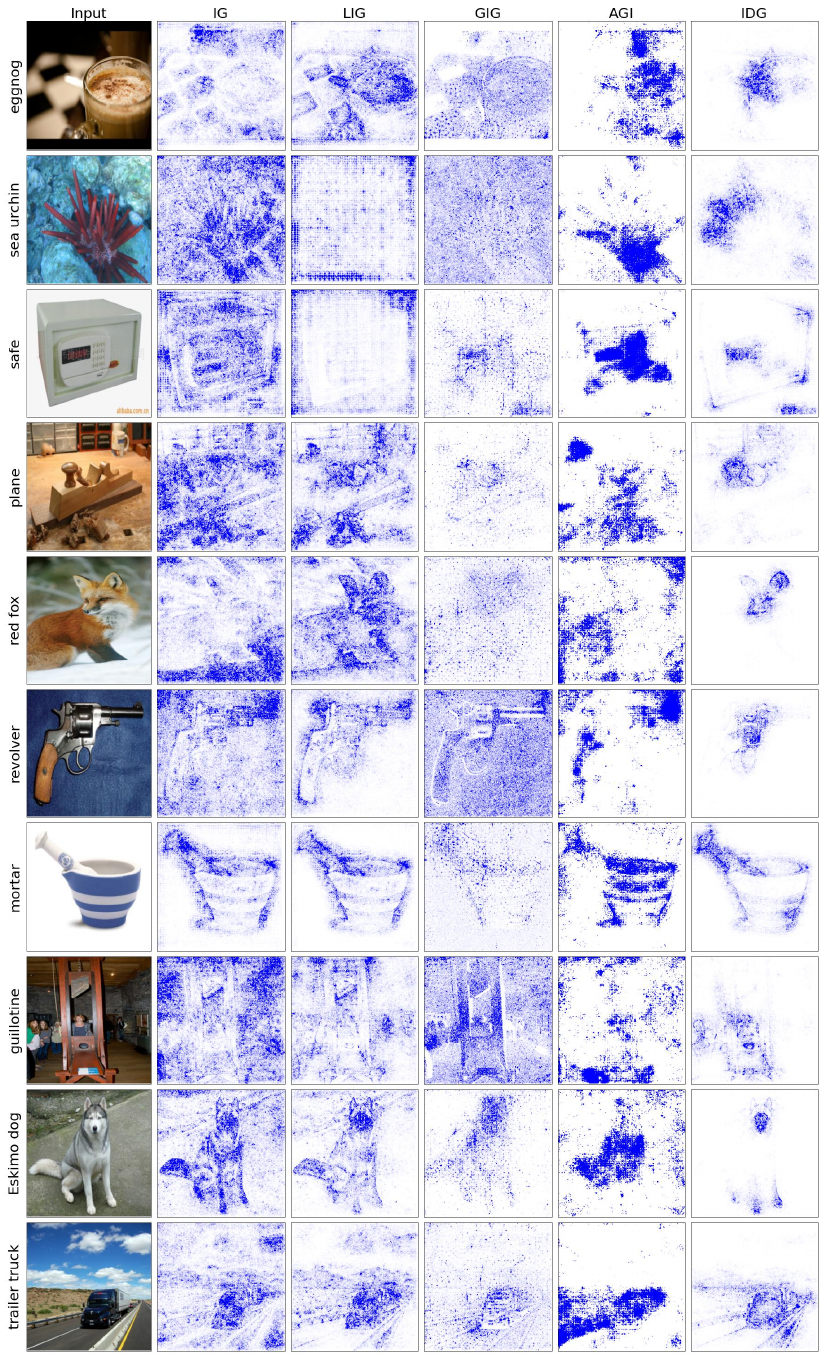}

\bibliography{aaai24}

\end{document}